\documentclass[10pt,twocolumn,letterpaper]{article}

\usepackage{cvpr}              

\usepackage{graphicx}
\usepackage{amsmath}
\usepackage{amssymb}
\usepackage{booktabs}
\usepackage{multirow}
\usepackage{amsfonts}
\usepackage{algorithm}
\usepackage{algorithmic}
\usepackage{bm}
\usepackage[dvipsnames]{xcolor}
\usepackage{colortbl}  
\usepackage{xcolor}

\usepackage[accsupp]{axessibility}

\usepackage[pagebackref=true,breaklinks=true,letterpaper=true,colorlinks,citecolor=ForestGreen, bookmarks=false]{hyperref}

\usepackage[capitalize]{cleveref}
\crefname{section}{Sec.}{Secs.}
\Crefname{section}{Section}{Sections}
\Crefname{table}{Table}{Tables}
\crefname{table}{Tab.}{Tabs.}

\def\confName{CVPR}
\def\confYear{2023}
\newcommand{\authorskip}{\hspace{5mm}}

\newcommand\blfootnote[1]{%
  \begingroup
  \renewcommand\thefootnote{}\footnote{#1}%
  \addtocounter{footnote}{-1}%
  \endgroup
}

\begin{document}

\title{SVFormer: Semi-supervised Video Transformer for Action Recognition}

\author{Zhen Xing\textsuperscript{1,2} \authorskip Qi Dai\textsuperscript{3} \authorskip Han Hu\textsuperscript{3} \authorskip Jingjing Chen\textsuperscript{1,2} \authorskip Zuxuan Wu\textsuperscript{1,2$\dagger$} \authorskip Yu-Gang Jiang\textsuperscript{1,2} \\[0.5mm]
{
\textsuperscript{1} Shanghai Key Lab of Intell. Info. Processing, School of CS, Fudan University} \\
{\textsuperscript{2} Shanghai Collaborative Innovation Center of Intelligent Visual Computing}
\\
{\textsuperscript{3} Microsoft Research Asia}
}
\maketitle

\begin{abstract}
\blfootnote{$^{\dagger}$ Corresponding author.}

Semi-supervised action recognition is a challenging but critical task due to the high cost of video annotations. Existing approaches mainly use convolutional neural networks, yet current revolutionary vision transformer models have been less explored. In this paper, we investigate the use of transformer models under the SSL setting for action recognition. 
To this end, we introduce SVFormer, which adopts a steady pseudo-labeling framework (\ie, EMA-Teacher) to cope with unlabeled video samples.
While a wide range of data augmentations have been shown effective for semi-supervised image classification, they generally produce limited results for video recognition.
We therefore introduce a novel augmentation strategy, Tube TokenMix, tailored for video data where video clips are mixed via a mask with consistent masked tokens over the temporal axis.
In addition, we propose a temporal warping augmentation to cover the complex temporal variation in videos, which stretches selected frames to various temporal durations in the clip.
Extensive experiments on three datasets Kinetics-400, UCF-101, and HMDB-51 verify the advantage of SVFormer. In particular, SVFormer outperforms the state-of-the-art by 31.5\% with fewer training epochs under the 1\% labeling rate of Kinetics-400. Our method can hopefully serve as a strong benchmark and encourage future search on semi-supervised action recognition with Transformer networks. Code is released at \url{https://github.com/ChenHsing/SVFormer}.

\end{abstract}

\section{Introduction}
\label{sec:intro}

\begin{figure}[t]

  \centering
   \includegraphics[width=1.0 \linewidth]{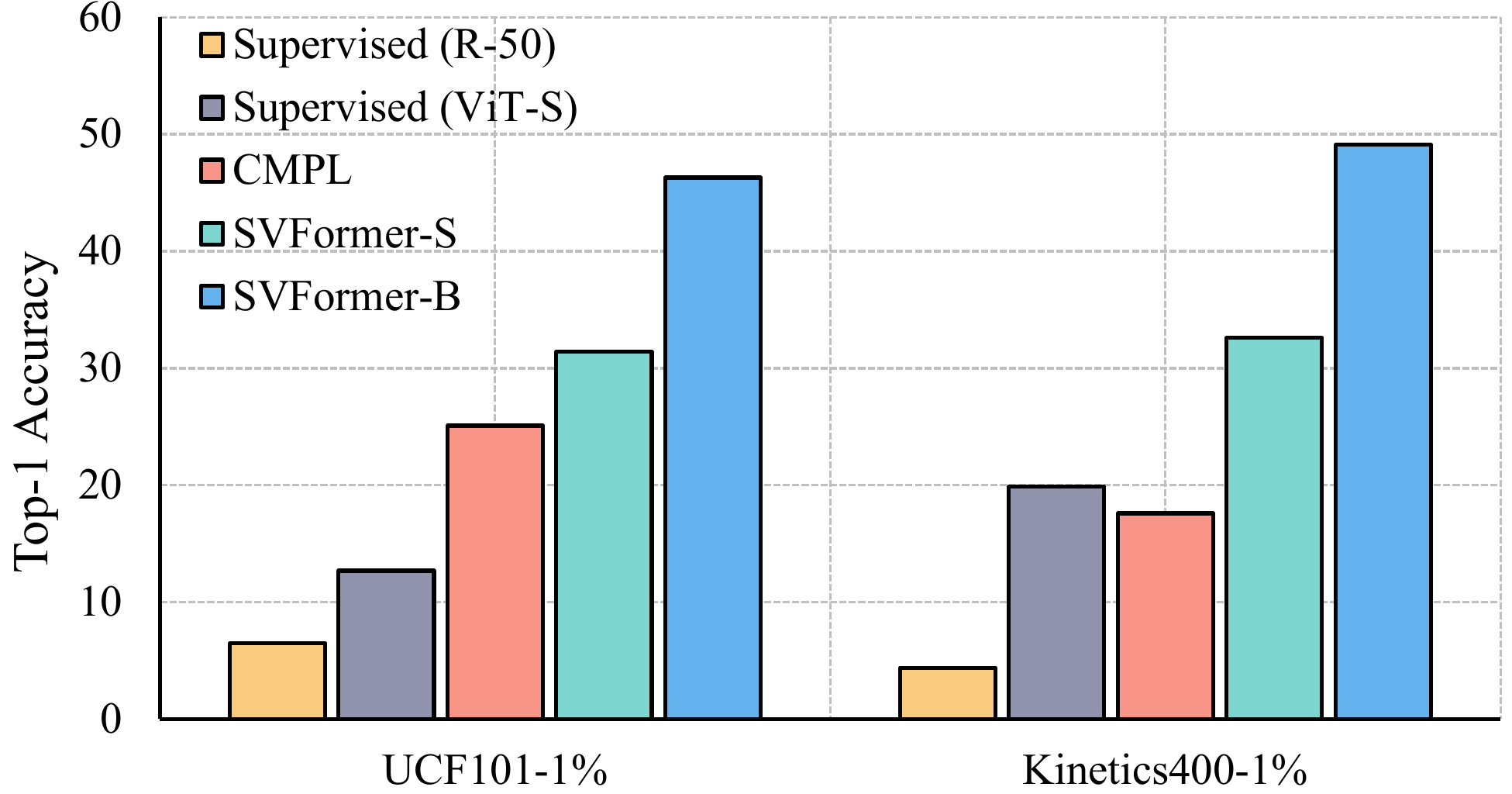}
   \vspace{-0.25in}
   \caption{Comparison of our method with the supervised baseline and previous state-of-the-art SSL method \cite{cmpl}. SVFormer significantly outperforms previous methods under the case with very little labeled data.}
   \label{bar}
   \vspace{-0.25in}

\end{figure}

Videos have gradually replaced images and texts on Internet and grown at an exponential rate. On video websites such as YouTube, millions of new videos are uploaded every day. Supervised video understanding works \cite{slowfast,x3d,timesformer,videoswin, coarse2fine,videosum, zhu2020actbert} have achieved great successes. They rely on large-scale manual annotations, yet labeling so many videos is time-consuming and labor-intensive. How to make use of unlabeled videos that are readily available for better video understanding is of great importance~\cite{videossl,shi2020weakly,shi2021temporal}.

In this spirit, semi-supervised action recognition \cite{videossl,tcl,ltg} explores how to enhance the performance of deep learning models using large-scale unlabeled data. This is generally done with labeled data to pretrain the networks~\cite{ltg,cmpl}, and then leveraging the pretrained models to generate pseudo labels for unlabeled data, a process known as pseudo labeling. The obtained pseudo labels are further used to refine the pretrained models. 
In order to improve the quality of pseudo labeling, previous methods \cite{ltg,mvpl} use additional modalities such as optical flow \cite{opticalflow} and temporal gradient \cite{tsn}, or introduce auxiliary networks \cite{cmpl} to supervise unlabeled data. 
Though these methods present promising results, they typically require additional training or inference cost, preventing them from scaling up.

Recently, video transformers \cite{videoswin,timesformer,vivit} have shown strong results compared to CNNs \cite{x3d,slowfast,3dresnet}.
Though great success has been achieved, the exploration of transformers on semi-supervised video tasks remains unexplored.
While it sounds appealing to extend vision transformers directly to SSL, a previous study shows that transformers perform significantly worse compared to CNNs in the low-data regime due to the lack of inductive bias~\cite{semiformer}. As a result, directly applying SSL methods, \eg, FixMatch \cite{fixmatch}, to ViT \cite{vit} leads to an inferior performance~\cite{semiformer}.

Surprisingly, in the video domain, we observe that TimeSformer, a popular video Transformer \cite{timesformer}, initialized with weights from ImageNet \cite{imagenet}, demonstrates promising results even when annotations are limited~\cite{rahman2022surprising}.
This encourages us to explore the great potential of transformers for action recognition in the SSL setting.

Existing SSL methods generally use image augmentations (\eg, Mixup \cite{mixup} and CutMix \cite{cutmix}) to speed up convergence under limited label resources.
However, such pixel-level mixing strategies are not perfectly suitable for transformer architectures, which operate on tokens produced by patch splitting layers.
In addition, strategies like Mixup and CutMix are particularly designed for image tasks, which fail to consider the temporal nature of video data.
Therefore, as will be shown empirically, directly using Mixup or CutMix for semi-supervised action recognition leads to unsatisfactory performance.

In this work, we propose SVFormer, a transformer-based semi-supervised action recognition method.
Concretely, SVFormer adopts a consistency loss that builds two differently augmented views and demands consistent predictions between them.
Most importantly, we propose Tube TokenMix (TTMix), an augmentation method that is naturally suitable for video Transformer.
Unlike Mixup and CutMix, Tube TokenMix combines features at the token-level after tokenization via a mask, where the mask has consistent masked tokens over the temporal axis.
Such a design could better model the temporal correlations between tokens.

Temporal augmentations in literatures (\emph{e.g.} varying frame rates) only consider simple temporal scaling or shifting, neglecting the complex temporal changes of each part in human action.
To help the model learn strong temporal dynamics, we further introduce the Temporal Warping Augmentation (TWAug), which arbitrarily changes the temporal length of each frame in the clip. TWAug can cover the complex temporal variation in videos and is complementary to spatial augmentations \cite{autoaugment}.
When combining TWAug with TTMix, significant improvements are achieved.

As shown in Fig.~\ref{bar}, SVFormer achieves promising results in several benchmarks. (i) We observe that the supervised Transformer baseline is much better than the Conv-based method \cite{3dresnet}, and is even comparable with the 3D-ResNet state-of-the-art method \cite{cmpl} on Kinetics400 when trained with 1\% of labels. (ii) SVFormer-S significantly outperforms previous state-of-the-arts with similar parameters and inference cost, measured by FLOPs. (iii) Our method is also effective for the larger SVFormer-B model.
Our contributions are as follows:
\begin{itemize}

\item We are the first to explore the transformer model for semi-supervised video recognition. Unlike SSL for image recognition with transformers, we find that using parameters pretrained on ImageNet is of great importance to ensure decent results for action recognition in the low-data regime.
\item We propose a token-level augmentation Tube TokenMix, which is more suitable for video Transformer than pixel-level mixing strategies. 
Coupled with Temporal Warping Augmentation, which improves temporal variations between frames, TTMix achieves significant boost compared with  image augmentation.
\item We conduct extensive experiments on three benchmark datasets. The performances of our method in two different sizes (\emph{i.e.}, SVFormer-B and SVFormer-S) outperform state-of-the-art approaches by clear margins. Our method sets a strong baseline for future transformer-based works.
\end{itemize}

\section{Related Works}
\noindent\textbf{Deep Semi-supervised Learning}~
Deep learning relies on large-scale annotated data, however collecting these annotations is labor-intensive.
Semi-supervised learning is a natural solution to reduce the cost of labeling, which leverages a few labeled samples and a large amount of unlabeled samples to train the model. 
The research and application of SSL mainly focus on image recognition \cite{fixmatch,semiformer,meanteacher} with a two-step process: data augmentation and consistency regularization. Concretely, different data augmentations \cite{autoaugment}  views are input to the model, and their output consistencies are enforced through a consistency loss. Another line of work generates new data and labels using mixing~\cite{mixmatch,ict,cowmask} to train the network. Among these state-of-the-art methods, FixMatch have been widely used due its effective and its variants have been extended to many other applications, such as object detection \cite{unbiasedteacher,rethinkingsemi}, semantic segmentation \cite{semiseg1,semiseg2}, 3D reconstruction \cite{ssl3d}, etc. Although FixMatch has achieved good performance in many tasks, it may not achieve satisfactory results when directly transferred to video action recognition due to the lack of temporal augmentation. In this paper, we introduce temporal augmentation TWAug with mixing based method TTMix, which is suitable for video transformers at SSL settings.

\vspace{0.1cm}
\noindent\textbf{Semi-supervised Action Recognition}
VideoSSL \cite{videossl} presents a comparative study of applying 2D SSL methods to videos, which verifies the limitations of the direct extension of pseudo labeling method. TCL \cite{tcl} explores the effect of a group contrastive loss and self-supervised tasks. MvPL \cite{mvpl} and LTG \cite{ltg} introduce optical flow or temporal gradient modal to generate high quality pseudo labels for training, respectively. CMPL \cite{cmpl} introduce an auxiliary network, which requires more frames in training, increasing the difficulty of application. Besides, previous methods are all based on 2D \cite{tcl} or 3D convolutional \cite{mvpl,cmpl,ltg} networks, which require more training epochs. Our approach is the first to make the exploration of Video Transformer for SSL action recognition and achieves the best performance with the least training cost.

\noindent\textbf{Video Transformer}
The great success of vision transformer \cite{vit,liu2021swin,zhanghivit,tian2023resformer, panoswin} in image recognition leads to the development of exploring the transformer-base architecture for video recognition tasks. VTN \cite{vtn} uses additional temporal attention on the top of the pretrained ViT \cite{vit}. TimeSformer \cite{timesformer} investigates different spatial-temporal attention mechanisms and adopts factored space time attention as a trade-off of speed and accuracy. ViViT \cite{vivit}  explores four different types of attention, and selects the global spatio-temporal attention as the default to achieve promising performance. In addition, MviT \cite{mvit}, Video Swin \cite{videoswin} , Uniformer \cite{uniformer} and Video Mobile-Former~\cite{videomobile}  incorporate the inductive bias in convolution into  transformers. While these methods focus on fully-supervised setting,  limited effort has been made for  transformers in the semi-supervised setting.

\vspace{0.2cm}
\noindent\textbf{Data Augmentation}~
Data augmentation is an essential step in modern deep networks to improve the training efficiency and performance.
Cutout \cite{devries2017improved} removes random rectangle regions in images.
Mixup \cite{mixup} performs image mixing by linearly interpolating both the raw image and labels.
In CutMix \cite{cutmix}, patches are cut and pasted among image pairs. 
AutoAugment \cite{autoaugment} automatically searches for augmentation strategies to improve the results.
PixMix \cite{pixmix} explores the natural structural complexity of images when performing mixing.
TokenMix \cite{liu2022tokenmix} mixes images at the token-level and allows the region to be multiple isolated parts.
Though these methods have achieved good results, they are all specially designed for pure image and most of them are pixel-level augmentations.
In contrast, our TTMix coupled with TWAug is devised for video.

\section{Method}

In this section, we first introduce the preliminaries of SSL in Sec.~\ref{preliminary}. 
The pipeline of our proposed SVFormer is described in Sec.~\ref{pipeline}.
Then we detail the proposed Tube TokenMix (TTMix) in Sec.~\ref{tokenmix}, as well as the effective albeit simple temporal warping augmentation.
Finally, we show the training paradigm in Sec.~\ref{train}.

\subsection{Preliminaries of SSL}
\label{preliminary}
Suppose we have $N$ training video samples, including $N_L$ labeled videos $(x_l,y_l) \in \mathcal{D}_L$ and $N_U$ unlabeled videos $x_u \in \mathcal{D}_U$, where $x_l$ is the labeled video sample with a category label $y_l$, and $x_u$ is the unlabeled video sample. In general, $N_U \gg N_L$. The aim of SSL is to utilize both $\mathcal{D}_L$ and $\mathcal{D}_U$ to train the model.

\subsection{Pipeline}
\label{pipeline}

SVFormer follows the popular semi-supervised learning framework FixMatch \cite{fixmatch} that use a consistency loss between two differently augmented views. The training paradigm is divided into two parts. 
For the labeled set $\{(x_l, y_l)\}_{l=1}^{N_L}$, the model optimizes the supervised loss ${\cal{L}}_{s}$:
\begin{equation}
\label{losssup}
{\cal{L}}_{s}=\frac{1}{N_L}\sum_{}^{N_L}{\cal{H}}({\cal{F}} ({x_l}),y_l),
\end{equation}
where ${\cal{F}}(\cdot)$ refers to the predictions produced by the model and ${\cal{H}}$ is the standard cross entropy loss. 

For unlabeled samples $x_u$, we first use weak augmentations (\emph{e.g.}, random horizontal flipping, random scaling, and random cropping) and strong augmentations (\emph{e.g.}, AutoAugment~\cite{autoaugment} or Dropout~\cite{dropout}) to generate two views separately, $x_w={\cal{A}}_{weak}(x_u)$, $x_s={\cal{A}}_{strong}(x_u)$. Then the pseudo label of the weak view ${\hat{y}_w}=\arg\max({\cal{F}}(x_w))$, which is produced by the model, is utilized to supervise the strong view, with the following unsupervised loss:
\begin{equation}
\label{lossun}
{\cal{L}}_{un}=\frac{1}{N_U}\sum^{N_U}{\mathbb{I}} (\max({\cal{F}}(x_w)) > \delta){\cal{H}}({\cal{F}}({x_s}),\hat{y}_w),
\end{equation}
where $\delta$ is the predefined threshold, and ${\mathbb{I}}$ is the indicator function that equals $1$ when the maximum class probability exceeds $\delta$ otherwise $0$. The confidence indicator is used to filter the noisy pseudo labels.

\begin{figure}[t]
  \centering
   \includegraphics[width=1 \linewidth]{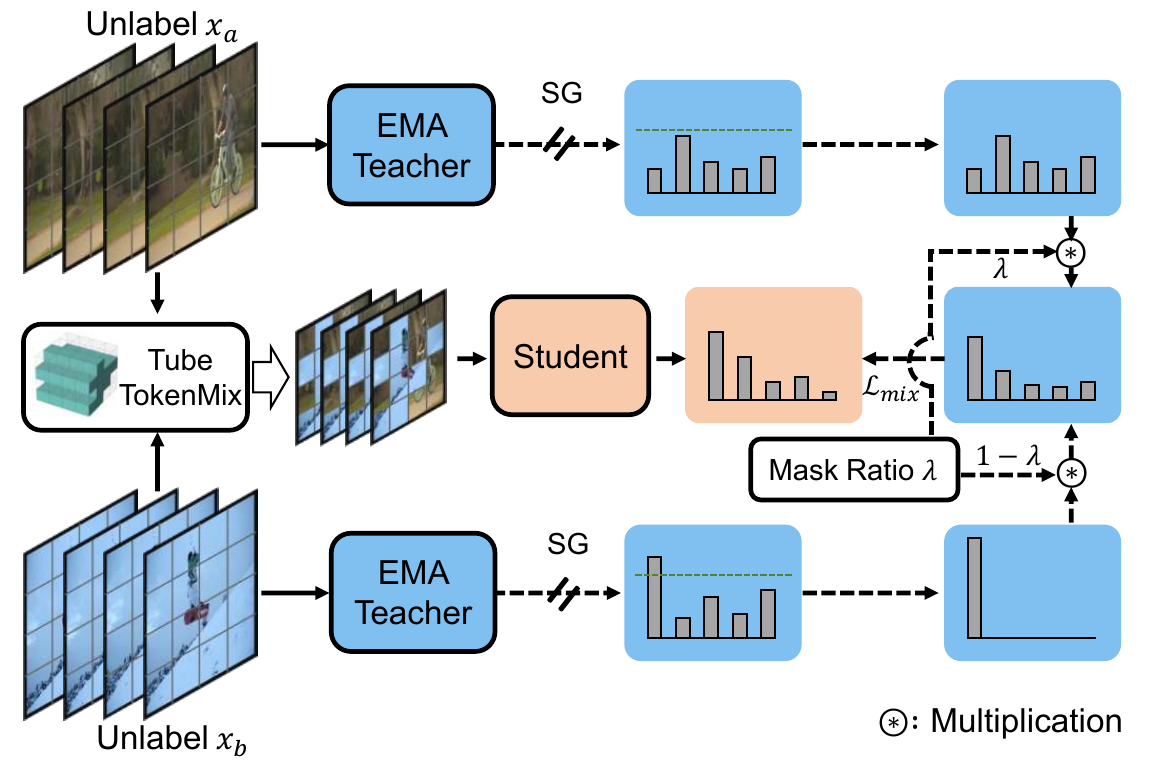}
      \vspace{-0.25in}
   \caption{ Overview of our Tube TokenMix training framework. The two input unlabeled samples are mixed via a tube mask, where the generated sample is fed into the Student model.
   The two samples are also fed into the EMA-Teacher to obtain their pseudo labels, which are further linearly interpolated via mask ratio $\lambda$ to produce pseudo label for the generated sample.
``SG" means stop gradient.
 The data augmentation is omitted here.}
   \label{overview}
\end{figure}

\paragraph{EMA-Teacher}
In FixMatch, the two augmented inputs share the same model, which tends to cause model collapsing easily \cite{moco,bootstrap}. 
Therefore, we adopt the exponential moving average (EMA)-Teacher in our framework, which is an improved version of FixMatch.
The pseudo labels are generated by the EMA-Teacher model, whose parameters are updated by exponential moving average of the student parameters, formulated as:
\begin{equation}
\label{ema}
{\theta}_t \gets m {\theta}_t +(1- m) {\theta}_s,
\end{equation}
where $m$ is a momentum coefficient, $\theta_{t}$ and $\theta_{s}$ are the parameters of teacher and student model, respectively. EMA has achieved success in many  tasks, such as self-supervised learning \cite{moco,bootstrap}, SSL of image classification \cite{meanteacher,cowmask}, and object detection \cite{unbiasedteacher,rethinkingsemi}. Here we are the first to adopt this method in semi-supervised video action recognition.

\subsection{Tube TokenMix}
\label{tokenmix}

One of the core problems in semi-supervised frameworks is how to enrich the dataset with high-quality pseudo labels. 
Mixup \cite{mixup} is a widely adopted data augmentation strategy, which performs convex combination between pairs of samples and labels as follows:
\begin{equation}
  \hat{x} = \lambda \cdot x_1 + (1-\lambda) \cdot x_2,
\end{equation}
\begin{equation}
  \hat{y} = \lambda \cdot y_1 + (1-\lambda) \cdot y_2,
\end{equation}
where the ratio $\lambda$ is a scalar that conforms to the beta distribution. 
Mixup~\cite{mixup} and its variants (\emph{e.g.} CutMix \cite{cutmix}) have achieved success in many tasks in the low-data regime, such as long-tail classification \cite{remix,videolt}, domain adaptation \cite{damiup} \cite{dualmixup}, few-shot learning \cite{mixupfew,mpcn}, \etc.
For SSL, Mixup also performs well by mixing the pseudo labels of unlabeled samples in image classification \cite{ict}.

\paragraph{Mixing in Videos}
While directly applying Mixup or CutMix to video scenarios results in clear improvements in Conv-based methods~\cite{actorcutmix}, these methods show unsatisfactory performance in our method.
The reason is that our method adopts the Transformer backbone, where the pixel-level mixing augmentation (Mixup or CutMix) may be not perfectly suitable for such token-level models~\cite{liu2022tokenmix}.
To narrow the gap, we propose 3 token-level mixing augmentation methods for video data, namely, Rand TokenMix, Frame TokenMix, and Tube TokenMix.

\begin{figure}[t]
  \centering
   \includegraphics[width=1.0 \linewidth]{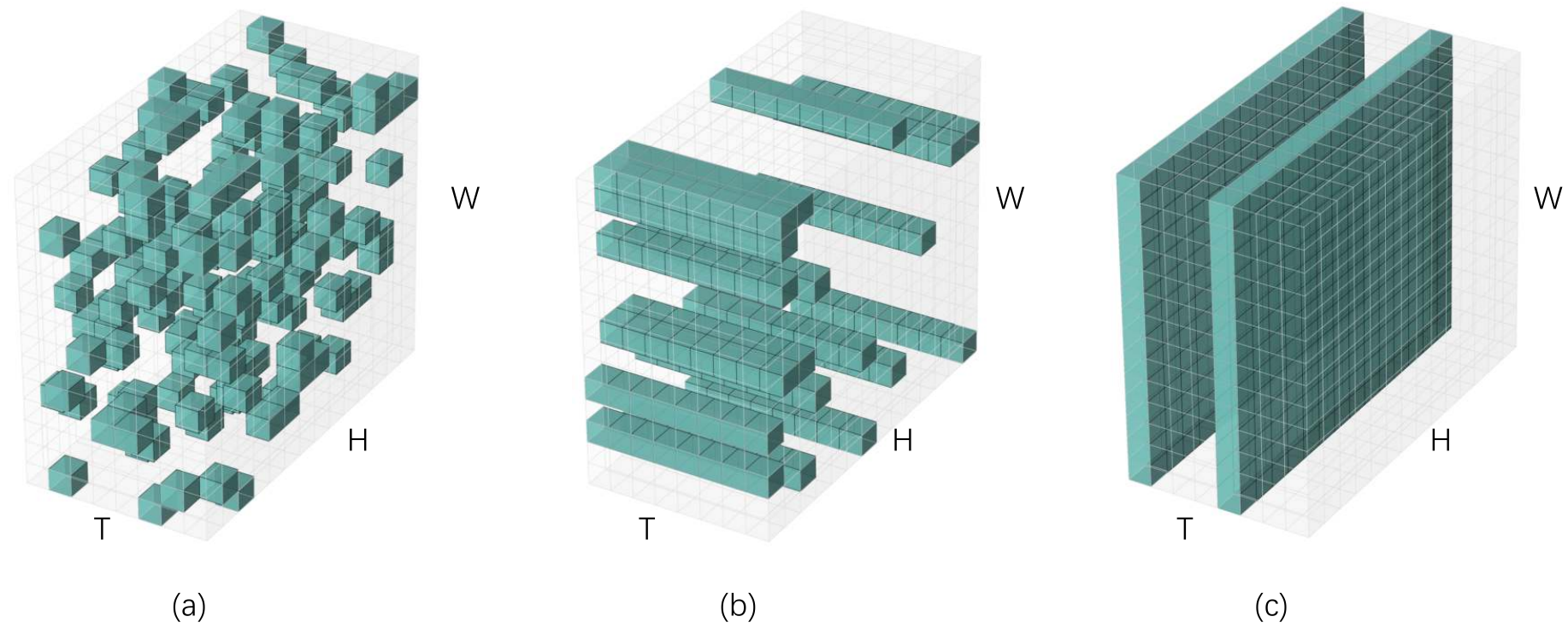}
      \vspace{-0.25in}
   \caption{Examples of masks in three token mixing strategies. (a) Rand TokenMix. (b) Tube TokenMix. (c) Frame TokenMix.}
   \label{mask3}
\end{figure}

Fig.~\ref{overview} illustrates the pipeline of our method.
Given unlabeled video clips $x_a, x_b\in \mathbb{R}^{H\times W\times T}$, our method employ a token-level mask $\textbf{M}\in\{0,1\}^{H\times W\times T}$ to perform sample mixing.
Note that $H$ and $W$ are the height and width of the frame after patch tokenization, and $T$ is the clip length. To generate a new sample $x_{mix}$, we mix $x_a$ and $x_b$ after strong data augmentations ${\cal{A}}_{strong}$ as follows:
\begin{equation}
    x_{mix} = {\cal{A}}_{strong}(x_a) \odot \textbf{M} + {\cal{A}}_{strong}(x_b) \odot (\textbf{1}-\textbf{M}),
\end{equation}
where $\odot$ is element-wise multiplication, and $\textbf{1}$ is a binary mask with all ones.

The mask $\textbf{M}$ differs in the three augmentation methods, as demonstrated in Fig.~\ref{mask3}. 
For Rand TokenMix, the masked tokens are randomly selected from the whole video clip (from $H\times W\times T$ tokens).
For Frame TokenMix, we randomly select frames from the $T$ frames and mask all the tokens in these frames.
For Tube TokenMix, we adopt the tube-style masking strategy, that is, different frames share the same spatial mask matrix. In this case, the mask $\textbf{M}$ has consistent masked tokens over the temporal axis.
While our mask design shares similarity with the recent masked image/video modeling \cite{feichtenhofer2022masked,tong2022videomae,mvd,bevt,xie2022simmim,xie2022data}, our motivation is totally different. They focus on removing certain regions and making the model predict the masked areas for feature learning.
In contrast, we leverage the mask to mix two clips and synthesize a new data sample.

\begin{algorithm}[t]
\caption{Consistency loss for Tube TokenMix}
\begin{algorithmic}
\REQUIRE Unlabeled clip batch $\mathbf{x_a}$
\REQUIRE Tube TokenMask $\mathbf{M}$, mask ratio $\lambda$
\REQUIRE Teacher model ${\cal{F}}_{t}$
\REQUIRE Student model ${\cal{F}}_{s}$
\REQUIRE Confidence threshold $\delta$
\STATE $\mathbf{{x}_{b}} = \text{shuffle}(\mathbf{x_a})$ \COMMENT{shuffle samples in batch}
\STATE $\mathbf{\hat{x}_a} = \text{spatial\_aug}(\mathbf{x_a})$ \COMMENT{spatial aug.}
\STATE $\mathbf{\hat{x}_b} = \text{temporal\_aug}(\mathbf{{x_b}})$ \COMMENT{temporal warping aug.}
\STATE $\mathbf{\hat{y}}_{a} = \text{stop\_gradient}({\cal{F}}_{t}(\mathbf{{x}}_{a}))$ \COMMENT{teacher pred.}
\STATE $\mathbf{\hat{y}}_{b} = \text{stop\_gradient}({\cal{F}}_{t}(\mathbf{{x}}_{b}))$ 
\STATE $\mathbf{{c}_a} = \max_{i}\mathbf{y}_{a}[i]$   \COMMENT{confidence of prediction}
\STATE $\mathbf{{c}_b} = \max_{i}\mathbf{y}_{b}[i]$ 
\STATE $\mathbf{x}_{mix} = \mathbf{\hat{x}}_{a} * \mathbf{M} + \mathbf{\hat{x}}_{b} * (1 - \mathbf{M})$ \COMMENT{mix clips}
\STATE $\mathbf{\hat{y}}_{mix} = \mathbf{\hat{y}}_{a} * \lambda + \mathbf{\hat{y}}_{b} * (1- \lambda)$ \COMMENT{mix tea. preds.}
\STATE $\mathbf{c}_{mix} =\mathbf{c}_a * \lambda + \mathbf{c}_b *(1-\lambda) $ \COMMENT{mix confidence.}
\STATE $\mathbf{q} = \text{mean}(c_{mix} \geq \delta)$   \COMMENT{mean of conf. mask}
\STATE $\mathbf{y}_{mix} = {\cal{F}}_{s}(\mathbf{x}_{mix})$ \COMMENT{stu. pred. on mixed clip}
\STATE $ {\cal{L}}_{mix} = q||\mathbf{y}_{mix} - \mathbf{\hat{y}}_{mix}||^{2}_{2}$   \COMMENT{cons. loss}
\STATE \textbf{Return} $ {\cal{L}}_{mix}$
\end{algorithmic}
\label{alg}
\end{algorithm}

\begin{figure*}[t]

  \centering
   \includegraphics[width=1.0 \linewidth]{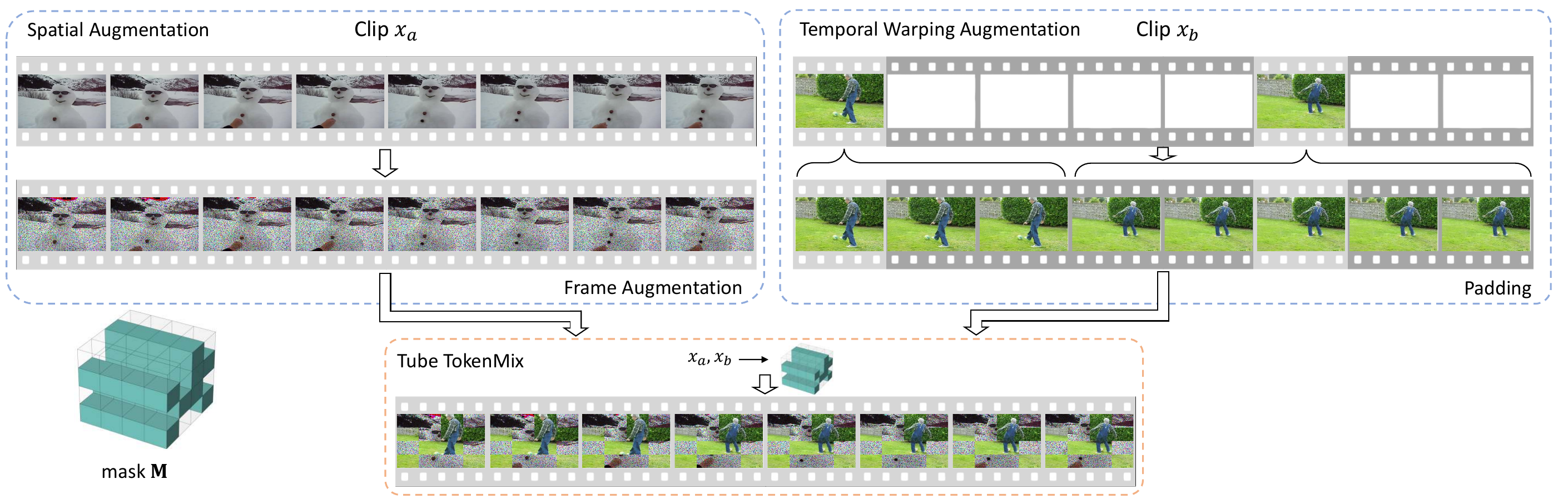}
      \vspace{-0.25in}
   \caption{One example of utilizing Temporal Warping Augmentation together with spatial augmentation in TTMix. The two clips are first transformed by the two augmentations separately, after which TTMix is performed to generate a new clip.}
   \label{augmix}
\end{figure*}

The mixed sample $x_{mix}$ is then fed to the student model ${\cal{F}}_{s}$, obtaining the model prediction $y_{mix} = {\cal{F}}_{s}(x_{mix})$.
In addition, the pseudo labels $\hat{y}_a,\hat{y}_b$ for $x_a,x_b$ are produced by inputting the weak augmented samples ${\cal{A}}_{weak}(x_a),{\cal{A}}_{weak}(x_b)$ to the teacher model ${\cal{F}}_{t}$:

\begin{equation}
    \hat{y}_a = \arg\max({\cal{F}}_{t}({\cal{A}}_{weak}(x_a))), 
\end{equation}
\begin{equation}
    \hat{y}_b = \arg\max({\cal{F}}_{t}({\cal{A}}_{weak}(x_b))).   
\end{equation}
Note that if $\small \max({\cal{F}}_{t}({\cal{A}}_{weak}(x)))<\delta$, the pseudo label $\hat{y}$ remains the soft label ${\cal{F}}_{t}({\cal{A}}_{weak}(x))$.
The pseudo label $\hat{y}_{mix}$ for $x_{mix}$ is generated by mixing $\hat{y}_a$ and $\hat{y}_b$ with mask ratio $\lambda$:
\begin{equation}
    \hat{y}_{mix} = \lambda \cdot \hat{y}_a + (1-\lambda) \cdot \hat{y}_b.
\end{equation}
Finally, the student model is optimized by the following consistency loss:
\begin{equation}
\label{lossmix}
    {\cal{L}}_{mix}= \frac{1}{N_{m}} \sum^{N_{m}}(\hat{y}_{mix} - y_{mix})^2,
\end{equation}
where $N_{m}$ is the number of mixed samples. 
The algorithm of consistency loss for TTMix is shown in Algorithm \ref{alg}.

\paragraph{Temporal Warping Augmentation}
\label{time_aug}
Most existing augmentation methods are designed for image tasks, which focus more on the spatial augmentation. They manipulate single or a pair of images to generate new image samples, without considering any temporal changes.
Even the commonly adopted temporal augmentations, including varying temporal locations \cite{feichtenhofer2021large} and frame rates \cite{tcl,yang2020video}, only consider simple temporal shift or scaling, that is, changing the holistic location or play speed.
However, human actions are very complex and can have different temporal variation at every timestamp.
To cover such challenging cases, we propose to distort the temporal duration of each frame, thus introducing higher randomness into the data.

\begin{figure}[h]

  \centering
   \includegraphics[width=1.0 \linewidth]{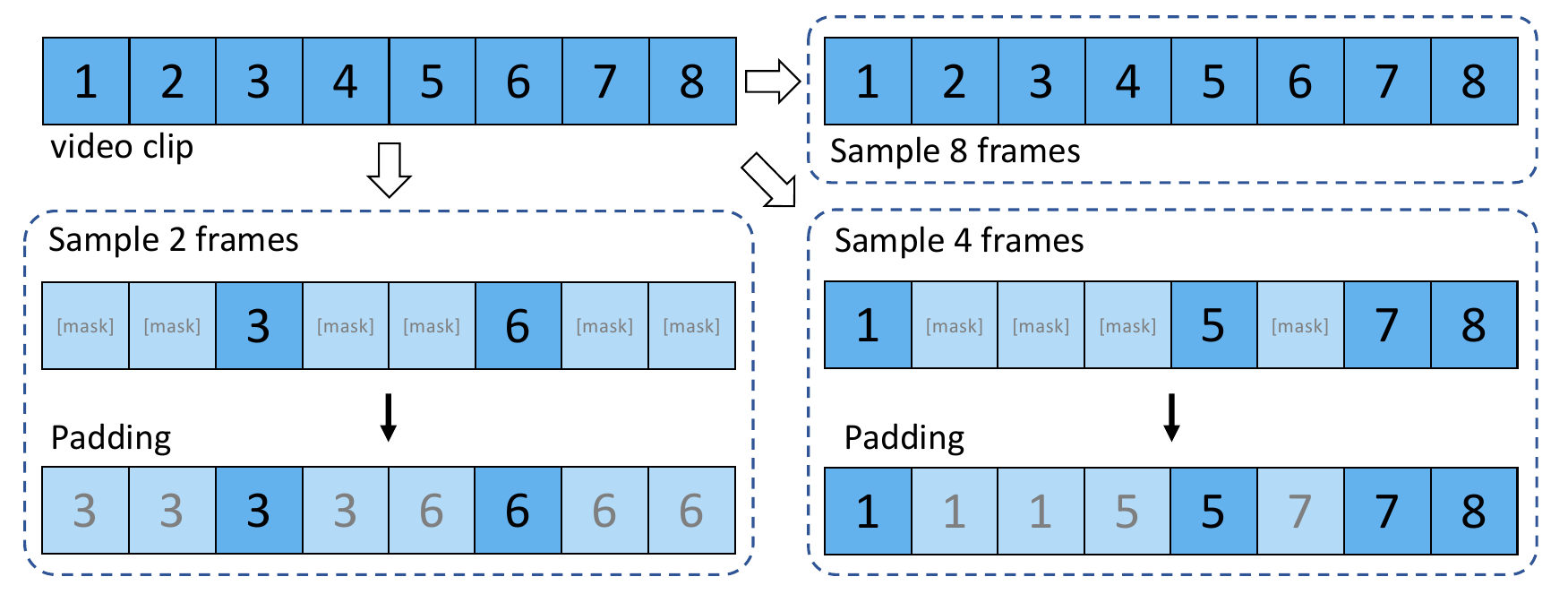}
   \vspace{-0.1in}
   \caption{Illustration of Temporal Warping Augmentation. We demonstrate three augment examples of different selected frames.}
   \label{fig_warp}
   \vspace{-0.2in}

\end{figure}

Our Temporal Warping Augmentation (TWAug) can stretch one frame to various temporal length.
Given an extracted video clip of $T$ frames (\emph{e.g.}, 8 frames), we randomly determine to keep all the frames, or select a small portion of frames (\emph{e.g.}, 2 or 4 frames) while masking the others.
The masked frames are then padded with random neighbouring visible (unmasked) frames.
Note that after temporal padding, the frame order is still retained.
Fig.~\ref{fig_warp} shows three examples of selecting 2, 4, and 8 frames, respectively.
The proposed TWAug can help the model learn the flexible temporal dynamics during training.

The Temporal Warping Augmentation serves as a strong augmentation in TTMix.
Typically, we combine TWAug with the conventional spatial augmentation \cite{autoaugment,dropout} to perform the mixing.
As shown in Fig.~\ref{augmix}, the two input clips are first transformed by spatial augmentation and TWAug separately, after which the two clips are mixed through TTMix.
We verify the effectiveness of our TWAug in Sec. \ref{augment}.

\subsection{Training Paradigm}
\label{train}
The training of SVFormer consists of three parts: supervised loss formulated by Eq. (\ref{losssup}), unsupervised pseudo-label consistency loss Eq. (\ref{lossun}), and TTMix consistency loss Eq. (\ref{lossmix}). The final loss function is as follows:
\begin{equation}
\label{losstotal}
{\cal{L}}_{all}= {\cal{L}}_{s} + {\gamma}_1  {\cal{L}}_{un} + {\gamma}_2  {\cal{L}}_{mix},
\end{equation}
where ${\gamma}_1$ and ${\gamma}_2$ are the hyperparameters for balancing the loss items. 

\begin{table*}[t]
\centering
\caption{\textbf{Comparisons with state-of-the-art methods on UCF-101 and Kinetics-400.} Note that 3D-ResNet-18 and 3D-ResNet-50 denote the backbone
networks and their depths. We report Top-1 accuracy as the evaluation metric. ``Input'' shows the modalities used during training, where ``V'' is the raw RGB video, ``F'' is optical flow and ``G'' is the temporal gradient.
} 
 \vspace{-0.0in}

\scalebox{0.97}{
\begin{tabular}{lcccccccc}
\toprule
\multirow{2}{*}{Method}     & \multirow{2}{*}{Backbone} & \multirow{2}{*}{Input} & \multirow{2}{*}{w ImgNet} & \multirow{2}{*}{Epoch}  & \multicolumn{2}{c}{UCF-101} & \multicolumn{2}{c}{Kinetics-400} \\ \cline{6-9} 
                            &                                                   &             &   &        & 1\%          & 10\%         & 1\%             & 10\%           \\ \cline{1-9}
\multirow{2}{*}{Supervised} & 3D-ResNet-50              & V  & \checkmark                    & 200                   & 6.5          & 32.4         & 4.4             & 36.2           \\


                            & ViT-S                     & V   &\checkmark                   & 30                       & 12.7            & 62.5            & 19.9            & 56.6           \\

                            \midrule
FixMatch (NeurIPS 2020) \cite{fixmatch}     & SlowFast-R50              &V  &\checkmark           &200       &16.1 &55.1    &10.1          &49.4   \\
VideoSSL(WACV 2021) \cite{videossl}         & 3D-ResNet-18              & V  &\checkmark                     & -                      & -            & 42.0           & -               & 33.8              \\
TCL (CVPR 2021) \cite{tcl}             & TSM-ResNet-18                   & V     &                 & 400                    & -            & -            & 8.5             & -              \\
ActorCutMix (CVIU 2021) \cite{actorcutmix}    & R(2+1)D-34                & V    &\checkmark                  & 600                     & -            & 53.0         & 9.02            & -              \\
MvPL (ICCV 2021)\cite{mvpl}            & 3D-ResNet-50              & V+F+G     &              & 600                   & 22.8         & 80.5         & 17.0            & 58.2           \\
CMPL (CVPR 2022) \cite{cmpl}           & R50 + R50-1/4              & V   &\checkmark                    & 200                    & 25.1         & 79.1         & 17.6            & 58.4           \\
LTG (CVPR 2022) \cite{ltg}             & 3D-ResNet-18              & V+G     &                & 180/360                & -            & 62.4         & 9.8             & 43.8           \\
TACL(TCSVT 2022) \cite{tacl} &3D-ResNet-50  &V &\checkmark  & 200   &-  & 55.6  &- &- \\
L2A (ECCV 2022) \cite{l2a}    &3D-ResNet-18               &V  &\checkmark   &400 &- &60.1
&- &-  \\ 
\midrule

SVFormer-S (Ours)                        & ViT-S                     & V    &\checkmark                   & 30                   & 31.4            & 79.1            & 32.6           & 61.6           \\
SVFormer-B (Ours)                        & ViT-B                     & V  &\checkmark                    & \textbf{30}                    & \textbf{46.3}           & \textbf{86.7}            & \textbf{49.1}      & \textbf{69.4}           \\ \bottomrule

\end{tabular}
}
\label{table1}
\end{table*}

\section{Experiment}
In this section, we first introduce the experimental settings in Sec.~\ref{experimentsettings}. Following previous work \cite{cmpl,ltg}, we conduct experiments under different labeling rates in Sec \ref{mainresult}. In addition, we also perform ablation experiments and empirical analysis in Section \ref{ablation}. If not emphasized, we only use RGB modal for inference with the official validation set.

\subsection{Experiment Settings}
\label{experimentsettings}
\noindent\textbf{Datasets}
Kinetics-400 \cite{kinetic} is a large-scale human action video dataset, with up to 245k training samples and 20k validation samples, covering 400 different categories. We follow the state-of-the-art methods MvPL \cite{mvpl} and CMPL \cite{cmpl} to sample 6 or 60 labeled training videos per category, \emph{i.e.} at 1\% or 10\% labeling rates. UCF-101 \cite{ucf101} is a dataset with 13,320 video samples, which consists of 101 categories. We also sample 1 or 10 samples in each category as the labeled set following CMPL \cite{cmpl}. As for HMDB-51 \cite{hmdb}, it is a small-scale dataset with only 51 categories composed of 6,766 videos. Following the division of LTG \cite{ltg} and VideoSSL \cite{videossl}, we conduct experiments at three different labeling rates: 40\%, 50\%, and 60\%.

\vspace{0.05in}
\noindent\textbf{Evaluation Metric}
We show the accuracy of Top-1 in main results, and also present the accuracy of Top-5 in some ablation experiments.

\vspace{0.05in}
\noindent\textbf{Baseline}
We utilize the ViT \cite{vit} extended video TimeSformer \cite{timesformer} as the backbone of our baseline. The hyperparameters are mostly kept the same as the baseline, and we adopt the divided space-time attention as in TimeSformer \cite{timesformer}. Since TimeSformer only have ViT-Base models, we implement SVFormer-Small model from DeiT-S \cite{deit} with the dimension of 384 and 6 heads, in order to have comparable number of parameters with other Conv-based methods \cite{3dresnet,mvpl,cmpl}.
For fair comparisons, we train 30 epochs for TimeSformer as the supervised baseline.

\begin{table}[t]
\caption{\textbf{Comparisons with state-of-the-art methods on HMDB-51.}  We report Top-1 accuracy.  ``Input'' shows the modalities used during training, where ``V'' is the raw RGB video, ``F'' is optical flow and ``G'' is the temporal gradient.
}
\vspace{-0.0in}
\scalebox{0.85}{
\begin{tabular}{@{}cccccc@{}}
\toprule
            & Backbone   &Input  & 40\% & 50\% & 60\% \\ \midrule
VideoSSL \cite{videossl}    & 3D-R18     & V    & 32.7 & 36.2 & 37.0 \\
ActorCutMix\cite{actorcutmix}  & R(2+1)D-34 & V    & 32.9 & 38.2 & 38.9 \\
MvPL \cite{mvpl}        & 3D-R18     & V+F+G    & 30.5 & 33.9 & 35.8 \\
LTG \cite{ltg}        & 3D-R18     & V+G    & 46.5 & 48.4 & 49.7 \\
TACL \cite{tacl}        & 3D-R18     & V    & 38.7 & 40.2 & 41.7 \\ 
L2A \cite{l2a}        & 3D-R18     & V    &42.1   & 46.3 &47.1 \\
\midrule
SVFormer-S (Ours)        & ViT-S      & V      & \textbf{56.2}  &\textbf{58.2}    & \textbf{59.7}     \\ 
SVFormer-B (Ours)        & ViT-B      & V      & \textbf{61.6}  &\textbf{64.4}    & \textbf{68.2}     \\ 
\bottomrule
\end{tabular}
}
\label{table2}
\vspace{-0.2in}

\end{table}

\vspace{0.00in}
\noindent\textbf{Training and Inference}
For training, we follow the setting of TimeSformer \cite{timesformer}. The training uses 8 or 16 GPUs, with a SGD optimizer using a momentum of 0.9 and a weight decay of 0.001.  For each setting, the basic learning rate is set to 0.005, which is divided by 10 at epochs 25, and 28. As for the confidence score threshold, we search for the optimal $\delta$ from \{0.3, 0.5, 0.7, 0.9\}. ${\gamma}_1$ and ${\gamma}_2$ are set to 2. 
The masking ratio $\lambda$ is sampled from beta distribution Beta($\alpha, \alpha$), where $\alpha=10$.
In the testing phase, following the inference strategies in MvPL \cite{mvpl} and CMPL \cite{cmpl}, we uniformly sample five clips from the entire video, and make three different crops to get $224\times224$ resolution to cover most of the spatial areas of the clips. The final prediction is the average of the softmax probabilities of these $5\times3$ predictions. We also conduct a comparison of inference setting in the ablation study in Sec. \ref{augment}.

\subsection{Main Results}
\label{mainresult}
The main results of Kinetics-400 \cite{kinetic} and UCF-101 \cite{ucf101} are shown in Table \ref{table1}. Compared with previous methods, our model SVFormer-S achieves the best performance with the fewest training epochs among the methods that only use RGB data. In particular, at the labeling rate of 1\% setting, SVFormer-S improves previous approach~\cite{cmpl} by 6.3\% in UCF-101 and 15.0\% in Kinetics-400. 
In addition, when adopting larger models, SVFormer-B significantly outperforms the state-of-the-art methods.

Specifically, in Kinetics-400, SVFormer-B can achieve 69.4\% with only 10\% labeled data, which is comparable to 77.9\% of fully-supervised setting in TimeSformer \cite{timesformer}. 
Moreover, as shown in Table \ref{table2}, for the small-scale dataset HMDB-51 \cite{hmdb}, our SVFormer-S and SVFormer-B have also improved by about 10\% and 15\% compared with the previous method \cite{ltg}.

\begin{figure}[t]
  \centering
   \includegraphics[width=1.0 \linewidth]{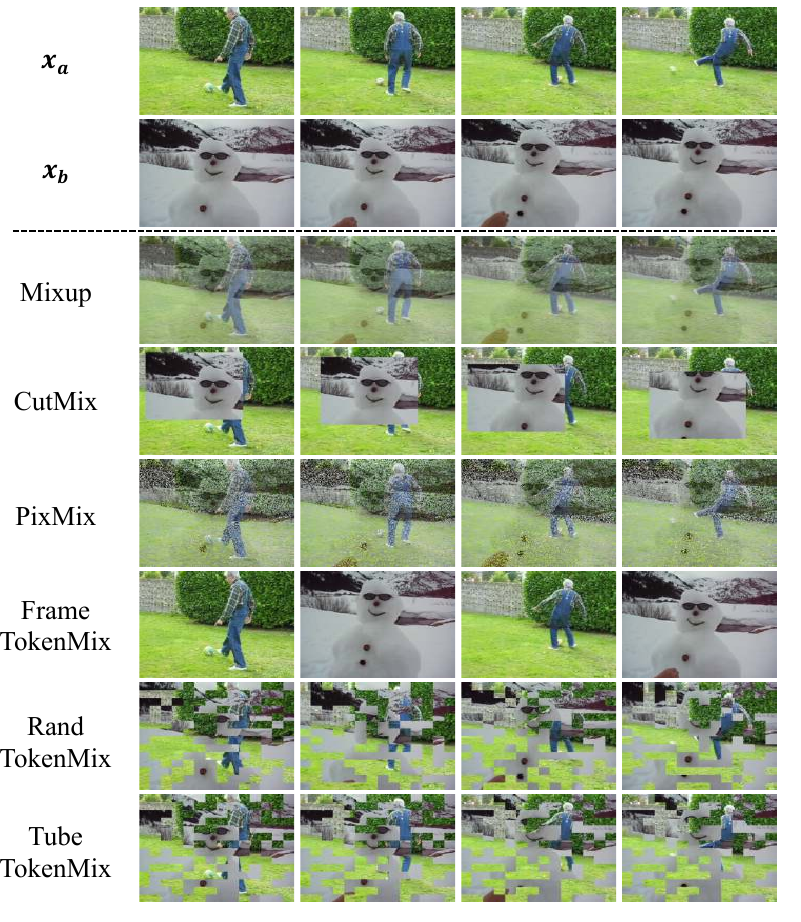}
   
   \caption{ Example of the traditional pixel-level mixing methods and our proposed token-level mixing. Note that token-level methods mix two samples after tokenization but the visualization is shown on image-level for clear presentation.}
   \label{fig_mixexample}
   \vspace{-0.5cm}
\end{figure}

\subsection{Ablation Studies}
\label{ablation}
To understand the effect of each part of the design in our method, we conduct extensive ablation studies on the Kinetics-400 and UCF-101 at the 1\% labeling ratio setting with SVFormer-S.

\vspace{0.05in}
\noindent\textbf{Analysis of SSL framework}
The comparison of FixMatch and EMA-Teacher is shown in Table \ref{table3}. 
It is clear that the two methods have significantly improved the baseline approach.
In addition, EMA-Teacher has exhibited considerable gains over FixMatch in both datasets with very few labeled samples, probably because it has improved the stability of training. FixMatch \cite{fixmatch} may lead to model collapse with limited labels as shown in \cite{semivit}.

\begin{table}[t]\small
\centering
\caption{\textbf{SSL framework selection.} We compare the EMA-Teacher framework with FixMatch. The results are reported on Kinetics-400 and UCF-101 with 1\% labeling ratio.}
\vspace{-0.0in}
\begin{tabular}{@{}lcccc@{}}
\toprule
\multirow{2}{*}{Method} & \multicolumn{2}{c}{UCF-1\%} & \multicolumn{2}{c}{Kinetic-1\%} \\ \cmidrule(l){2-5} 
                  & Top-1          & Top-5          & Top-1        & Top-5        \\ \midrule
Baseline         & 12.7             & 29.8          & 19.9         & 42.3   \\ \midrule
FixMatch \cite{fixmatch}          &  25.1              & 47.3               &  28.2       & 54.6             \\
 EMA-Teacher       & \textbf{31.4}       & \textbf{56.9}               & \textbf{32.6}    & \textbf{59.0}             \\ \bottomrule

\end{tabular}
\label{table3}
\end{table}

\begin{table}[t]\small
\centering
\caption{\textbf{TokenMix Mask sampling.}  We compare different token masking strategies. The results are reported on Kinetics-400 and UCF-101 with 1\% labeling ratio. }
\vspace{-0.1in}
\begin{tabular}{@{}lcccc@{}}
\toprule
\multirow{2}{*}{Method/Dataset} & \multicolumn{2}{c}{UCF-1\%}
& \multicolumn{2}{c}{Kinetic-1\%}  \\ \cmidrule(l){2-5} 
                                & Top-1          & Top-5          & Top-1        & Top-5        \\ \midrule
Baseline   &26.1 &48.9  &23.6 &47.7  \\                                
CutMix \cite{cutmix}             & 28.7            & 51.3                & 28.6           & 53.7                    \\
Mixup \cite{mixup}                & 29.8            & 53.0            & 29.3           & 55.1                      \\
PixMix \cite{pixmix}              & 29.7            & 52.4             & 29.6           & 55.8                     \\ \midrule
Frame TokenMix            & 29.8            & 54.2        & 26.3           & 50.0                      \\ 
Rand TokenMix          & 30.3            & 55.3            & 28.8           & 54.2                     \\ 
 Tube TokenMix            & \textbf{31.4}            & \textbf{56.9}         & \textbf{32.6}           & \textbf{59.0}                      \\ \bottomrule
\end{tabular}
\label{table4}
\vspace{-0.1in}

\end{table}

\vspace{0.05in}
\noindent\textbf{Analysis of different mixing strategies} 
We now compare the Tube TokenMix strategy with three pixel-level mixing methods, CutMix \cite{cutmix}, Mixup \cite{mixup}, PixMix \cite{pixmix}, as well as the other two token-level mixing methods, \emph{i.e.}, Frame TokenMix and Rand TokenMix. The examples of different mixing methods are shown in Fig. \ref{fig_mixexample}. The quantitative results are shown in Table \ref{table4}. Compared with these alternative methods, all mixing methods can improve the performance, which proves the effectiveness of mixing-based consistency losses.
In addition, we observe that the token-level methods (Rand TokenMix and Tube TokenMix) perform better than the pixel-level mixing methods.
This is not surprising since transformers operate on tokens, and thus token-level mixing has inherent advantages.

The performance of Frame TokenMix is even worse than that of pixel-level mixing methods, which is also expected.
We hypothesize that replacing entire frames in video clip will scramble up the temporal reasoning, thus leading to poor temporal attention modeling.
In addition, Tube TokenMix achieves the best results.
We suppose the consistent masked tokens over temporal axis can prevent information leaky between adjacent frames in the same spatial locations, especially in such a short-term clip. 
Therefore, Tube TokenMix could better model the spatio-temporal correlations.

\vspace{0.05in}
\noindent\textbf{Analysis of Augmentations}
\label{augment}
The effects of spatial augmentation and temporal warping augmentation are evaluated in Table \ref{table5}. 
The baseline indicates removing both strong spatial augmentation (\emph{e.g.} AutoAugment \cite{autoaugment} and Dropout \cite{dropout}) and TWAug in all branches.
In this case, the experimental performance drops dramatically. 
When spatial augmentation or temporal warping augmentation is incorporated into the baseline separately, the performance is improved. The best practice is to perform data augmentations in both spatial and temporal.

\begin{table}[t]
\centering
\caption{\textbf{Effects of Spatial and Temporal Warping Augmentations (TWAug).} The results are reported on Kinetics-400 and UCF-101  with 1\% labeling ratio.}
\vspace{-0.1in}

\scalebox{0.75}{
\begin{tabular}{@{}lcccccc@{}}
\toprule
\multirow{2}{*}{} &
\multirow{2}{*}{Spatial} &
\multirow{2}{*}{Temporal} &

\multicolumn{2}{c}{UCF-1\%} & \multicolumn{2}{c}{Kinetic-1\%} \\ \cmidrule(l){4-7} 
                 &  &  & Top-1 & Top-5  & Top-1  &Top-5 \\ \midrule
Baseline         &         &     &29.5 &52.1     & 28.5      & 54.2       \\ \midrule
Spatial-only     & \checkmark        &     & 29.9 & 53.2      & 30.3      & 55.9    \\
Temporal-only    &         &  \checkmark    & 30.2  & 55.8     & 30.8      & 56.4     \\
 Spatial+Temporal & \checkmark         &   \checkmark   & \textbf{31.4} &  \textbf{56.9}       &\textbf{32.6}       & \textbf{59.0}     \\ \bottomrule
\end{tabular}
}
\label{table5}
\end{table}

\begin{table}[t]
\centering
\caption{\textbf{Effects of different inference schemes.} We compare the sparse sampling strategy and dense sampling method with different frames. 
The $a\times b$ in `Frames' column means sampling $a$ frames at frame rate $b$.
The results are reported on Kinetics-400 and UCF-101 with 1\% labeling ratio.}
\vspace{-0.1in}
\scalebox{0.9}{
\begin{tabular}{@{}lcccc@{}}
\toprule
Method    & Frames    & Test View &UCF-1\% &Kinetic-1\%  \\ \midrule
MvPL \cite{mvpl}  & $8\times8$        &$10\times3$ & 22.8  & 17.0  \\
CMPL \cite{cmpl} & $8\times8$         & $10\times3$ & 25.1  & 17.6   \\ \midrule
SVFormer-S &~~$8\times32$         & ~~\bm{$1\times3$}  & 29.3 & 31.0    \\
SVFormer-S  & $8\times8$          & ~~$5\times3$  &31.4  &32.6     \\ 
SVFormer-S & $16\times4$~~          & $10\times3$   & \textbf{31.6}  & \textbf{33.1} \\ 

\bottomrule
\end{tabular}
}
\label{table6}
\vspace{-0.4cm}
\end{table}

\vspace{0.05in}
\noindent\textbf{Analysis of Inference}
We evaluate the effect of frame rate sampling and different inference schemes, as shown in Table \ref{table6}. 
Previous methods, \emph{i.e.} CMPL \cite{cmpl} and MvPL \cite{mvpl}, utilize the clip-based sparse sampling, which samples 8 frames at frame rate 8 as a clip. For each video, 10 clips are sampled, where each clip is cropped 3 times according to different spatial positions. Finally, the predictions of $10\times3$ samples are averaged. 
TimeSformer \cite{timesformer} adopts the video-based sparse sampling, that is, 8 frames are sampled at frame rate 32 as the representation of the whole video. Then 3 different cropped views are used, namely $8\times32$.  Applying the video-based sparse sampling scheme as in TimeSformer \cite{timesformer} can reduce the inference cost, but the performance is worse than that of clip-based sampling. 
Experiments at sampling schemes of $8\times8$ and $16\times4$ demonstrate better performance.
For the trade-off between efficiency and accuracy, we use $8\times8$ as the default setting following CMPL \cite{cmpl} and MvPL \cite{mvpl}.

\begin{figure}[t]
  \centering
   \includegraphics[width=0.95 \linewidth]{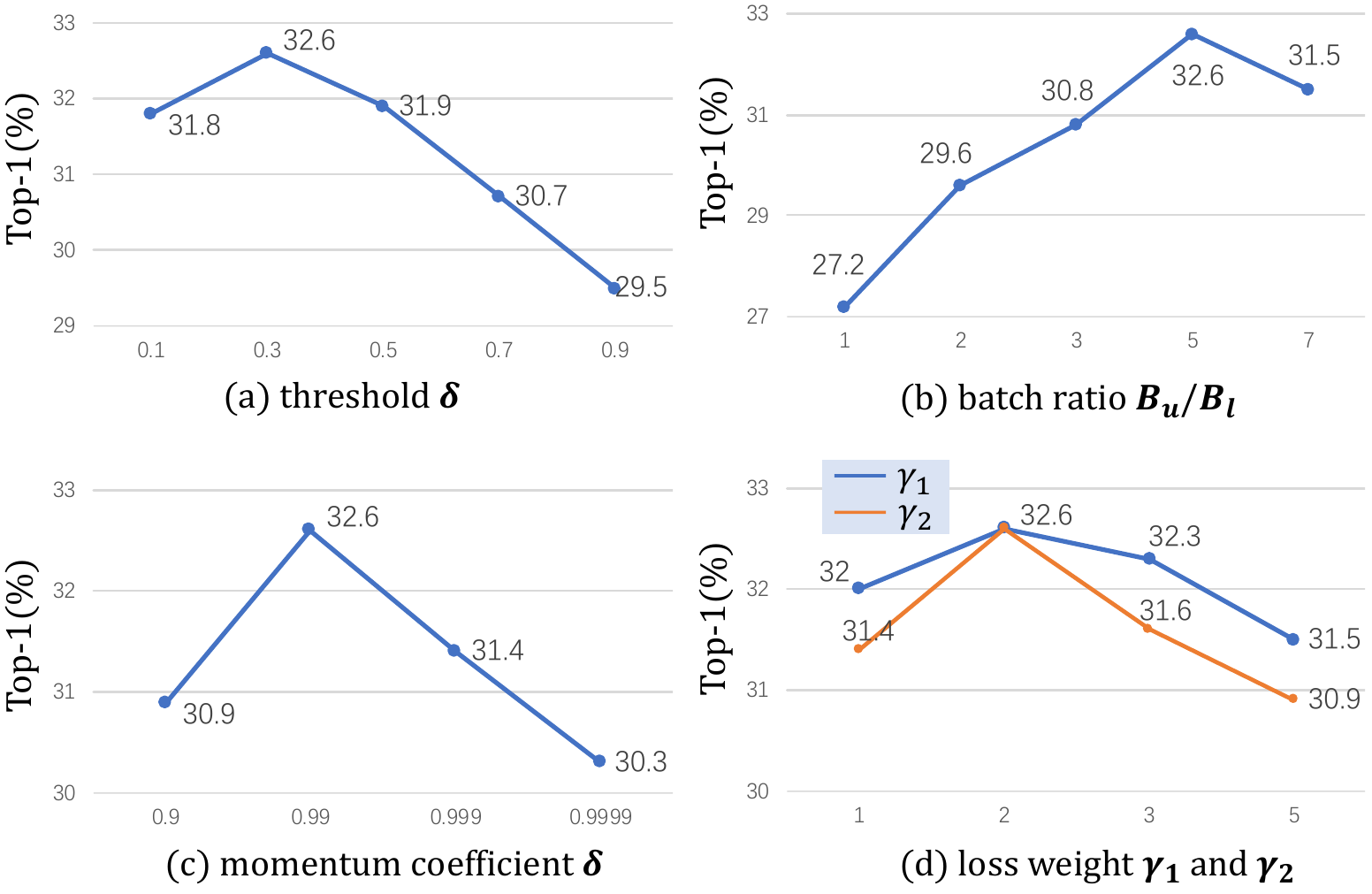}
       \vspace{-0.2cm}
   \caption{\textbf{Effects of hyperparameters.} Results of varying threshold $\delta$, ratio between unlabeled and labeled data in a mini-batch ($B_u/B_l$), momentum coefficient $m$, and loss weights (${\gamma}_1, {\gamma}_2$) are included to comprehensively study
the effects of the hyperparameters. Reported on Kinetics-400 with 1\% labeling ratio.}
   \label{fig7}
   \vspace{-0.5cm}
\end{figure}

\vspace{0.05in}
\noindent\textbf{Analysis of hyperparameters}
Here we explore the effect of different hyperparameters. We conduct experiments under 1\% setting of Kinetics-400.
We first explore the effect of different threshold values of $\delta$. 
As shown in Fig.~\ref{fig7}(a), we can observe that when labeled samples are extremely scarce, best results are achieved by setting a small $\delta$ value ($\delta=0.3$).
We then evaluate how the ratio between labeled samples and unlabeled samples in a mini-batch affect the result.
We fix the labeled sample number $B_l$ to 1, and sample $B_u$ unlabeled samples to form a mini-batch, where $B_u$ is in \{1, 2, 3, 5, 7\}.
The results are shown in Fig.~\ref{fig7}(b). When $B_u=5$, the model produces the highest result.
Finally, we explore the choice of momentum coefficient $m$ of EMA and the loss weights ${\gamma}_1$ and ${\gamma}_2$, as shown in Fig.~\ref{fig7}(c) and Fig.~\ref{fig7}(d). We thus set $m=0.99$ and ${\gamma}_1 = {\gamma}_2 = 2$ as default setting in all the experiments.

\section{Conclusion}

In this paper we present SVFormer, a transformer-based semi-supervised video action recognition method.
We propose Tube TokenMix, a data augmentation method that is specially designed for video transformer models.
Coupled with the temporal warping augmentation, which covers the complex temporal variations by arbitrarily changing the frame length, TTMix achieves significant improvement compared with conventional augmentations.
SVFormer outperforms the state-of-the-art with a large margin on UCF-101, HMDB-51 and Kinetics-400 without increasing overheads. Our work establishes a new benchmark for semi-supervised action recognition and encourages future work to adopt Transformer architecture.

\textbf{Acknowledgement} This project was supported by NSFC under Grant No. 62032006 and No. 62102092.

{\small
\bibliographystyle{ieee_fullname}
\bibliography{egbib}
}

\end{document}